# Innateness, AlphaZero, and Artificial Intelligence

Gary Marcus[1]

New York University

## Abstract

The concept of innateness is rarely discussed in the context of artificial intelligence. When it is discussed, or hinted at, it is often the context of trying to reduce the amount of innate machinery in a given system. In this paper, I consider as a test case a recent series of papers by Silver et al (Silver et al., 2017a) on AlphaGo and its successors that have been presented as an argument that a "even in the most challenging of domains: it is possible to train to superhuman level, without human examples or guidance", "starting *tabula rasa*."

I argue that these claims are overstated, for multiple reasons. I close by arguing that artificial intelligence needs greater attention to innateness, and I point to some proposals about what that innateness might look like.

[1] Departments of Psychology and Neural Science, New York University, gary.marcus at nyu.edu. This manuscript is based on a pair of lectures given at NIPS 2017, a brief conversation there with Demis Hassabis, and a debate that I had with Yann LeCun at NYU on October 5, 2017. I thank those audiences for discussion, and Dave Barner, Annie Duke, Ernie Davis, Pedro Domingos, Ken Forbus, Danny Kahneman, Stefano Pacifico, Ajay Patel, Elizabeth Spelke and Brad Wyble for comments.

# 1. An ancient debate, updated for the machine age

One of the oldest debates in intellectual history revolves around the somewhat nebulous[2] concept of innateness. How much of the human mind is built-in, and how much of it is constructed by experience? Plato famously become one of nativism's first advocates, in his dialogue the Meno, when he argued that a slave boy had knowledge of geometry, despite the lack of any formal training. In modern times, Noam Chomsky is perhaps the scholar most identified with nativism, having argued that human children could not acquire human language in the fashion that they (nearly) universally do, unless they were born with a "language acquisition device" (Chomsky, 1965), or what Steven Pinker (Pinker, 1994) has called a "language instinct." In contrast, empiricists, such as John Locke (Locke, 1694), and in modern times, the late developmental psychologist Elizabeth Bates and cognitive scientist Jeff Elman (Elman, 1996), have taken something close to a "blank slate" or *tabula rasa* view, arguing that our knowledge come from experience, delivered through the senses.

Virtually all modern observers would concede that genes and experience work together; it is "nature *and* nurture", not "nature versus nurture". No nativist, for instance, would doubt that we are also born with specific biological machinery that allows us to learn. Chomsky's Language Acquisition Device should be viewed precisely as an *innate* learning mechanism, and nativists such as Pinker, Peter Marler (Marler, 2004) and myself (Marcus, 2004) have frequently argued for a view in which a significant part of a creature's innate armamentarium consists not of specific knowledge but of learning mechanisms, a form of innateness that *enables* learning.

As discussed below, there is ample reason to believe that humans and many other creatures are born with significant amounts of innate machinery. The guiding question for the current paper is whether *artificially* intelligent systems ought similarly to be endowed with significant amounts of innate machinery, or whether, in virtue of the powerful learning systems that have recently been developed, it might suffice for such systems to work in a more bottom up, *tabula rasa* fashion.

# 2. Innateness in biological creatures

In a book length review of the biology of innateness (Marcus, 2004) I made a foundational distinction between what I called "prewiring" and "rewiring", and suggested that the biological evidence for both processes was overwhelming.

---

2 According to Mameli and Bateson (2006) the term innateness has been used in over two dozen ways, ranging from "universal" to "biological" to "acquired independently of learning"; it is the latter that I have in mind.

Over 90% of our genes are expressed in the development of the brain (Miller et al., 2014; Bakken et al., 2016; Kang et al., 2011), and a significant number of those are expressed selectively, in a way that allows the brain to self-assemble, even, to some non-trivial degree in the absence of experience. Mechanisms such as cell division, cell differentiation, cell migration, cell death, and axon guidance combine to self-assemble a rich first draft of the human brain, even prior to experience. Even in the absence of synaptic transmission, the primary mechanism by which experience is conveyed to the brain, the basic structure of the newborn brain is preserved (Verhage et al., 2000).

Commensurate with this, there is considerable evidence from the psychological literature that children are endowed early in life with what Spelke has called a "core knowledge" of domains like physics (Spelke & Kinzler, 2007). A long series of papers reviewed there suggests, for example, that infants have some ability to track and reason about objects. And, as Spelke (1994) has noted, it is difficult to see how knowledge of abstractions like objects, sets, and places could arise through (e.g.,) associative learning.

Vouloumanos and Werker (Vouloumanos and Werker, 2007a, 2007b) have shown that neonates can distinguish speech from closely-matched sine-wave analogues, even though the relevant information for making such distinctions is unlikely to percolate through to the womb. My own team discovered that 8-month-old infants could learn abstract rules from two minutes' exposure to artificial grammars (Gervain, Berent, & Werker, 2012). Gervain et al (2012) found that newborns could perform similar computations. A 2017 study in *Current Biology* which showed that late-term human fetuses could distinguish faces from inverted faces without any direct experience at all (Reid et al., 2017). Other work has suggested that deaf children can invent language with no direct model (Senghas, Kita, & Ozyürek, 2004)**,** and that language can be selectively impaired even in children with normal cognitive function (van der Lely & Pinker, 2014).

Elsewhere in the animal kingdom are many precocial animals with a capacity to walk and (to some degree) navigate obstacles within moments of birth. Newborn chicks appear to be able to recognize faces (Wood & Wood, 2015) and to distinguish biological motion from nonbiological motion (Mayer, Rosa-Salva, Morbioli, & Vallortigara, 2017).

Many vital questions remain. We don't know precisely what innate machinery humans (or other creatures) are born with, and we don't know how much of that machinery is tied to specific domains (e.g., language learning) and how much is domain-general (e.g., it is conceivable that mechanisms for acquiring representing abstract hierarchical structures might be useful in planning, motor control, language and other domains). And evidence for innateness doesn't mean that no (subsequent) learning takes place, it just means that whatever learning there is takes place against a background of some machinery that precedes learning; evidence for prewiring doesn't speak against rewiring, but nor does evidence for (later) rewiring speak against evidence for prewiring.

The bottom line is that there is more than enough evidence for innateness across multiple fields that we can reasonably presume for present purposes that humans (and other creatures) are born with significant amounts of innate machinery, and instead turn now to the related question of whether innate machinery might be a prerequisite for "human-level" artificial intelligence, sometimes referred to as "artificial general intelligence" [AGI]).

## 3.  Innateness in machines

One might think of cognition as a function over four variables

1. $cognition = f(a, r, k, e)$

where *a* = innate algorithms (whether domain-specific or domain-general), *r* = innate representational formats (again, domain-specific or otherwise), k = innate knowledge (again, domain-specific or otherwise), and e = experience. That is, the cognition of a given agent (biological or artificial) is a function of its innate algorithms, its innate representational formats, its innate knowledge, and experience.

A true blank slate would set *k* and *r* to zero, set *a* to some extremely minimal value (e.g., an operation for adjusting weights relative to reinforcement signals), and leave the rest to experience.  This is essentially the view that Locke (1694) articulated

> *All ideas come from sensation or reflection.*
>
> *Let us then suppose the mind to be, as we say, white paper, void of all characters, without any ideas:— How comes it to be furnished? ....*
>
> *Whence has it all the materials of reason and knowledge?*
>
> *To this I answer, in one word, from EXPERIENCE.*

Deep learning[3] pioneer Yann LeCun similarly appears to believe that both *a*, *r* and *k* should approach zero, as he made clear at an October 5, 2017 debate with me at NYU [tinyurl.com/lecunmarcusdebate]. In particular, in my own remarks I proposed a list of ten candidate elements that I felt would be important for AI, mostly on the representational side (see section 5). When questioned about that list by the moderator (David Chalmers) LeCun took a strong empiricist position, very much in the spirit of John Locke, suggesting that none of those 10 elements needed to be innate for AI systems. (Nor did he suggest any others.)

---

[3] For a recent critical appraisal of deep learning, see Marcus **{&, 2018, #96831}** and further discussion at https://medium.com/@GaryMarcus/in-defense-of-skepticism-about-deep-learning-6e8bfd5ae0f1

What AI researchers in general think about the values of *a*, *r*, *k*, and *e*, is largely unknown, in part because few researchers ever explicitly discuss the question. As Chalmers pointed out in a Facebook message to LeCun and myself in September 2017, nativism is hardly a central topic in contemporary AI thought. "When I do a google search on "nativism AI"", Chalmers wrote , "I get a "did you mean ..." message". Or, as Ken Forbus put it to me in an email, "the issue is literally not on the table for most AI people"; people get wrapped up in the practicalities of their immediate research, and don't really tend to think about nativism at all.

Others seem downright opposed, on principle; Tom Dietterich, for instance, told me in a recent email that "I think most ML people believe that methods for incorporating prior knowledge in the form of symbolic rules (or their probabilistic equivalent) are too heavy-handed and, while very useful from an engineering point of view, don't contribute to a plausible theory of general intelligence." (How to square this is the literature from biology, psychology, ethology, and neuroscience, I do not know.)

Some boundary conditions are clear. A classic result called the No Free Lunch theorem (Wolpert, 1996) effectively shows that *a* cannot literally be zero; every system will generalize in different ways, depending on what initial algorithm is specified, and no algorithm is uniquely best; different algorithms are suited to different problems. For example, as Pedro Domingos recently observed in an email

> *ML paradigms differ in is what assumptions they encode, and what form of additional knowledge they make it easy to encode. For example, neural nets assume continuity, graphical models assume conditional independence, and instance-based learning assumes similarity; and correspondingly, neural nets make it easy to incorporate types of continuity like translation invariance, graphical models [make it easy to incorporate] conditional independences, and [instance-based models make it easy to incorporate] knowledge of what makes things similar (in the kernel or distance measure, which will vary with the domain).*

Another classic paper, by Stuart Geman et al (Geman, Bienenstock, & Doursat, 1992)similarly showed that *r* can never be zero, and that the choice of *r* has significant impact on performance, arguably more important than the choice of algorithms. (Lachter and Bever(1988) and Pinker and Prince (Pinker & Prince, 1988) made this point more vividly but less formally, with respect to learning the English past tense.)

Perhaps the strongest argument for keeping the values of *a*, *r*, and *k* small, while relying on a high value of *e*, comes from DeepMind's groundbreaking work on playing classical board games through reinforcement learning, masterfully presented by Demis Hassabis at December 7, 2017 NIPS Symposium on Kinds of Intelligence, and in a series of three papers. The first, published in *Nature*, introduced AlphaGo (Silver et al., 2016); the second, also published in *Nature,* focused on a more powerful successor, AlphaGo Zero (Silver et al., 2017a); the third focused on Alpha Zero, a still-more powerful variation on the theme that played Go, chess and shogi at unprecedented levels, published on arXiv, December 2017 (Silver et al., 2017b).

Collectively, I will refer to these three systems, a marvel of AI engineering, as AlphaStar.

Before proceeding it's import clarify that although some people have read my recent critique of deep learning, and may imagine this paper to be extension of those, I don't view AlphaStar as a relatively unstructured "end-to-end" deep learning system, of the sort one typically finds in image recognition, in which one might expect nothing other than a deep network with many layers, with pixels on the input, and move choices on the output. Rather, AlphaStar is something much closer to the sort of thing I have been advocating: a deeply structured hybrid, making important use of deep learning, but also reliant on rich integration with more traditional symbolic techniques like tree search. It's a system in which deep learning is a fundamental tool, but embedded in a symbolic context. In that respect it is to some degree closer to what I was advocating architecturally.

And what AlphaStar does, it obviously does extremely effectively; my question here is not about whether it works, but about what the system's architecture and results *mean,* for thinking about nature and nurture with respect to AI.[4]

# 4. AlphaStar, and what it tells us about innateness

## 4.1. How DeepMind framed their results, and how they should actually be interpreted.

DeepMind's 2017 *Nature* paper frames their results, throughout, as an implicit, and at times explicit, argument for a strong version of empiricism. Their strongly antinativist framing began with title of the paper, which purported to show that they had demonstrated "Mastering the game of Go without human knowledge". The abstract similarly claimed that the system they presented achieved its undeniably impressive results by "starting *tabula rasa.*" The conclusions report that the paper had shown that "a pure reinforcement learning approach is fully feasible, even in the most challenging of domains: it is possible to train to superhuman level, without human examples or guidance, given no knowledge of the domain beyond basic rules", repeating the claim that system was able to do all this "starting *tabula rasa*."

---

[4] Another question, perhaps debatable, is whether AlphaStar should be considered to be an example of unsupervised learning. I think another (perhaps original?) term might be "self-supervised". Although AlphaGo Zero and AlphaZero don't require libraries of human games, they generate their own training databases, and learn from those in a partly supervised fashion. Also, where some unsupervised systems work entirely without labeled data, and induce essentially everything, AlphaGo Zero and Alpha Zero are given rules in advance and game-specific representations in advance, which is somewhat out of the spirit of traditional approaches to unsupervised learning. Either way, it is certainly an outstanding and effective example of reinforcement learning.

In reality, the system absolutely is *not* a *tabula rasa*, in the sense of being a literal blank slate; neither *a* nor *r* nor *k* in equation 1 are set to zero, as one might expect in a true blank slate. Nor is AlphaStar in fact a *pure* reinforcement learning system, or (contra the title) a system that altogether lacked human guidance.

Based on a reading of the abstract, a naive reader might presume, for example, that AlphaGo Zero might have induced — from experience — all that is needed to play in Go, and that the programmers themselves required no knowledge of Go in order to construct their system. The reality is quite different.

Reinforcement learning is supplement with other techniques, and human knowledge did in fact enter the system. Most of the paper's seventeen authors were deeply familiar with Go. One, Fan Hui, is a four time European go champion; ten were authors on the previous (2016) DeepMind paper that introduced AlphaGo, then the greatest Go program all time. In no way could the team be considered to be naive either to the nature of Go or the requirements necessary for building computer systems for computer Go.

Many aspects of their system follow both from previous studies of computer Go (and game playing in general) and from the nature of the problem itself. For instance, like virtually all computer Go systems of the last decade, the system built in Monte Carlo tree search, a technique, most often used in games, for evaluating moves and countermoves, with intermediate results accumulated and tested statistically over tree structures. Similarly, artfully placed convolutional layers allow the system to recognize that many patterns on the board are translation invariant.[5] (The convolutional layers are characteristic of deep learning systems, but because of the inclusion of the Monte Carlo Tree Search mechanisms, the systems as a whole is more of a hybrid, combining some aspects of deep learning with other aspects of classical symbol-manipulating systems.)

Crucially, the Monte Carlo tree structure apparatus was not *learned* from the data, by pure reinforcement learning. Rather, it was built in innately, into each iteration of AlphaStar, by DeepMind's programmers. (This tree search machinery was implemented as a mixture of algorithms and (tree) representations, hence contributed to both *a* and *r* not being zero in equation 1.)

---

[5] Groups of Go stones are not fully translational invariant, because of things like board edges and interactions with other clusters of stones; still a great deal of knowledge consists in knowing about patterns, some known as joseki, that are to, a first approximation, translationally invariant in their expect outcomes; I would be astonished if DeepMind could produce a variant of Alpha* that worked as well with comparable amounts of training time without incorporation convolution or something doing similar work, of recognizing common sets of patterns even as they appear in different locations across the game board.

Likewise, the convolutional layers[6] were structured in a precise way, not induced purely via reinforcement learning, with parameters appropriate for playing Go. AlphaGo Zero also included a special sampling algorithm for dealing with the symmetries (eg reflections and rotations) of Go boards, and both the board structure and the game rules themselves (representational assumptions, i.e., *r* in equation 1) were innate, rather than induced from visual images or exposure to games.

Crucially, these sampling mechanisms, arguably a mixture of innate algorithms and knowledge, were not included in AlphaZero's chess experiments, where presumably they would lead to a degradation, rather than an improvement (chess, unlike Go, is not symmetrical under rotation). This choice of which Monte Carlo augmentations to include or not include were again presumably made with human knowledge, rather than learned.

Interestingly, there is a strong case to be made that AlphaStar has far more innate machinery than DeepMind's earlier work on Atari games (Mnih et al., 2015) , which actually in at least four ways comes closer to being a genuinely blank slate. First, the Atari system did not preprogram in any game rules; second, it did not preprogram any game-specific representations (beyond the pixels of the display, the score, and the possible moves of an 8-direction joystick and accompanying "fire" button). Third, it did not include any game-specific data augmentation. Finally, fourth and most important, it did not build in any sort of tree search mechanisms.

Could the Atari system have been used for Go? Since it would have been so easy for them to try, I suspect that DeepMind tried applying their Atari game system to Go, but that the system failed, and that they did not report that failure. (If true, this deeply problematic. As Henderson et al (2017) have recently noted, a growing concern for the entire field of machine learning is that many things are tried, but few are published, with a vast array of hyperparameters, architectures, and representational schemes relegated to file drawers. Precisely this sort of replicability and file-drawer crisis has recently undermined psychology and medicine, and could lead to major problems within machine learning, too. The field of machine learning would greatly benefit from the sort of preregistration and systematic null result reporting in which those fields are now beginning to engage.)

On plausible assumption that the Atari system could not, unaltered, induce state-of-the-art Go performance, a reasonable inference is that AlphaGo's *superior performance stems in part from the additional innate structure that it embodies*, relative to the Atari game system.

Thus, rather being an illustration of the power of *tabula rasa* learning, AlphaGo is actually an illustration of the opposite: of the power of building in the right stuff to begin with.

---

[6]Convolutional layers build in what vision scientists call translational variance, allowing a system to recognize an object in any location as it moves across the image plane.

With the right initial algorithms and knowledge, complex problems are learnable (or learnable given some sort of real-world constraints on compute and data). Without the the right initial algorithms, representations and knowledge, many problems remain out of reach. Convolution is the prior that has made the field of deep learning work; tree search has been vital for game playing. AlphaZero has combined the two.

Ironies abound. To begin with, in some ways AlphaZero is actually more tied to certain kinds of innate representations and algorithms than people are. It is doubtful, for example, that humans have Monte Carlo tree search innately wired in and certainly not with the rules of Go or the structure of the board. To the extent that AlphaStar does build in innate algorithms, knowledge and representations, its constructs are *more* specific to Go and to game playing than any human might plausibly possess. Too much so, in fact; whereas a human can learn many games without specific innate representational features for any particular game, each implementation of AlphaStar is innately endowed with game-specific features that lock the system to one particular realization of one particular game, focusing on one particular problem. Humans are vastly more flexible in how they approach problems, and could, for example, answer a wide range questions about the game (is the current board symmetrical? Are there more black stones on the board than white? Where could I play if I wanted to deliberately lose territory?) or play on boards of a different size or shape, without retraining from scratch; AlphaStar could handle none of that. Most tellingly humans can quickly learn the rules and representations of novel games; AlphaStar at present has no capacity for learning such rules or game representations; instead it is entirely reliant on human programmers embedding those rules into the system innately.

The final irony is that nativists like Chomsky, Pinker, and myself have long argued that one of the most important starting conditions for learning language is having innate machinery for representing and manipulating trees. Rather than debunking our claims, DeepMind has provided fresh evidence for them, inasmuch as — despite considerable experimentation, with vast computational resources—they have been unable to succeed in complex board games without incorporating rich tree-theoretic innate starting points.

## 4.2 Beyond board games of perfect information

Go, chess, and shogi are often thought of as board games of perfect information; each player can, at any time, see exactly what is happening. There is no hidden information as in "fog of war", in which one player might be unaware of another maneuvers, and no uncertainty (as there is in poker or backgammon or the stock market). For this reason, games like Go, chess, and shogi are particularly amenable to a brute force, big data approach. A crucial question, in interpreting the AlphaStar results, is the degree to which the same mechanisms could solve a broader range of problems.

To a noticeable degree, the AlphaGo Zero paper invited the inference that their mechanisms might be general purpose in nature, both by referring to Go as "the most challenging domains" and by failing to consider any of the ways in which other problems (e.g., outside of Go) might not be amenable to the same sort of solution. Their conclusions in full read as follows, with little if any of the sort of hedging that is customary in scientific literature:

> Our results comprehensively demonstrate that a pure reinforcement learning approach is fully feasible, even in the most challenging of domains: it is possible to train to superhuman level, without human examples or guidance, given no knowledge of the domain beyond basic rules. Furthermore, a pure reinforcement learning approach requires just a few more hours to train, and achieves much better asymptotic performance, compared to training on human expert data. Using this approach, AlphaGo Zero defeated the strongest previous versions of AlphaGo, which were trained from human data using handcrafted features, by a large margin.
>
> Humankind has accumulated Go knowledge from millions of games played over thousands of years, collectively distilled into patterns, proverbs and books. In the space of a few days, starting *tabula rasa*, AlphaGo Zero was able to rediscover much of this Go knowledge, as well as novel strategies that provide new insights into the oldest of games."

No hedges or statements on the limits of scope were supplied, and the question of whether or not the result was general for other challenges was not addressed.

Alas, what's true for Go may not be true for many other challenges. Even relative to other board games, Go is challenging in some ways, but not all. Go demands strong pattern recognition and tree search skills, but other games are challenging in other ways. Civilization, for example, requires decision-making under uncertainty, design of transportation networks, and evaluating multiple kinds of tradeoffs (what to build, investment decisions, arms versus agriculture, etc). Diplomacy requires forming coalitions, demanding theory of mind. Charades requires acting skills, linguistic skills, and theory of mind, and so forth.

To take but one example, Moravčík et al's recent success on Poker (Moravčík et al., 2017), DeepStack, required a different (though related) set of innate structures, and it is no accident that the innate structures in DeepMind's Atari game experiments (Mnih et al., 2015) differ markedly from the innate machinery it used for perfect-information board games. The Atari games (as studied) were one-player games that could be largely played by the application of short-term tactics, hence tree search was of little value, whereas Go, chess and Shogi demand something like tree search because of the inherent profusion of alternatives. DeepMind tuned the machinery accordingly for each class of games. Even to handle the union of these two types of challenges (video games without prior knowledge, and board games with innate rules), DeepMind would likely require a broader set of innate machinery than either system used on its own.

And the further one goes from straightforward games, the more one may need to enrich the set of primitives. Ultimately, it seems likely that many different types of tasks will have their own innate requirements: Monte Carlo tree search for board games, syntactic tree manipulation operations for language understanding, geometric primitives for 3-D scene understanding, theory of mind for problems demanding social coalitions, and so forth.

Taken together, the full set of primitives may look less like a *tabula rasa* and more like the spatiotemporal manifold that Immanuel Kant (1781) envisioned, or like the sort of things that strong nativists like myself, Noam Chomsky, Elizabeth Spelke, Steve Pinker and the late Jerry Fodor have envisioned.

Unfortunately, DeepMind has not yet reported either success or failures of AlphaGo outside of the domain of perfect-information games, even on tasks such as Atari that they are quite familiar with. Although a naive reader might view AlphaZero's core technology as a kind of universal solvent, capable of solving a variety of challenge problems in a wide cross section of domains, that conjecture has not been explicitly tested in print.

Outside of Go, the prospects DeepMind has applied somewhat similar techniques (e.g., reinforcement learning, though not AlphaStar per se) to some other challenging problems, including language learning (Hermann et al., 2017). But the results there are not nearly as compelling as for Go or Chess; the AI systems are much slower to learn novel words than human children (in terms of amount of data required) and much slower to generalize semantic concepts like negation.

Even on alternative challenges *within the space of Go,* there is some reason for pessimism. For example, the system might need to be retrained from scratch in order to play on a game board of a different size or shape, or to play a game with a different goal (e.g. to lose rather than to win); it's unclear whether its knowledge could be repurposed towards teaching in the way that a human's knowledge could be. (Josh Tenenbaum, Elizabeth Spelke, and Ernie Davis all recently reminded of this point to me, in different ways, independently.) The grain level of what's innate in AlphaStar is a system that is built to play Go (or chess, or Shogi) on a board of a specific size; the grain level of what's innate in a person is utterly different; nothing like board sizes or rules are innate. There is also no evidence that the system could transfer what it has learned from one game to another, unlike humans who can to some extent transfer knowledge, e.g., between different turn-based-games.

In the final analysis, although the AlphaStar demonstrations are undeniably impressive, they simply cannot justify the broader claims that their 2017 *Nature* article seems to invite. Go is certainly a genuinely challenging problem, just as the DeepMind team suggests, but there is no particular reason to think that other hard problems would be solvable with the same innate machinery.

At a broader level, virtually every AI system contains lots of innate machinery that isn't acknowledged as such. Neural networks, for example, routinely contain innate assumptions about how many layers should be included, how many units should be in each layer, what the input and output units should stand for, what activation function individual units should follow, what sort of learning rule to use, what sort of learning rate to adopt, how many training examples should be used in a training sequence, and so forth; it has also become popular to use "curriculum-based learning", in which the (innate) habits of a supervisor or teacher are determined prior to learning. These commitments differ from one model to the next, e.g. LSTM's vs CNN's vs ResNets vs Memory networks vs Tree-RNNS, etc.. Only rarely are these commitments made however in terms of explicit discussions about innateness or qua hypotheses about putative cognitive functions.

All of this may be necessary in order to get systems to work, but too little of it seems *principled* with respect to the question of what should be innate, in part because virtually none of it engages with the sort of questions that a psychologist might ask about innateness, based on an understanding of how human cognitive creatures grow.

## 5. Innate machinery

If reinforcement learning and Monte Carlo tree search turn out not to be enough innate machinery, on their own, to support artificial general intelligence, what else might we be looking for? At my October 5, 2017 debate with Yann LeCun, I had an opportunity to draw up a preliminary list. The list I proposed was, roughly, the union of a set of computational primitives that I had advocated for in my book *The Algebraic Mind* (Marcus, 2001), and a set of conceptual primitives drawn from Elizabeth Spelke's work on cognitive development (Spelke, 1994):

- Representations of objects
- Structured, algebraic representations
- Operations over variables
- A type-token distinction
- A capacity to represent sets, locations, paths, trajectories, obstacles and enduring individuals
- A way of representing the affordances of objects
- Spatiotemporal contiguity
- Causality
- Translational invariance
- Capacity for cost-benefit analysis

Provocatively, LeCun argued (when pushed by the moderator, David Chalmers) that *none* of these need be innate.

As lavish my list of ten might seem to some researchers to be, in retrospect I suspect it is incomplete; one glaring omission is a representation of time, which I certainly should have included on the list. Intentionality (in the sense of inferring the intentions of others) perhaps should have been there, too. And maybe others. (Lake et al (Lake, Ullman, Tenenbaum, & Gershman, 2016)don't explicitly argue for nativism, and in fact explicitly avoid innate commitments, but point for a set of starting points for AI that are similar to those I suggest.)

With the right basis list, many other skills could of course be eventually acquired. Game-based tree-search (if I go there, and you go there, and then I go there, where will things stand?), for instance, may be innate in AlphaStar, but people probably *learn* how to do such analyses, albeit with less precision, putting together time, causality, and intentionality with a capacity for making cost-benefit analysis.

An interesting trend in the machine learning field is towards something one might call differentiable programming (Bošnjak, Rocktäschel, Naradowsky, & Riedel, 2016; Denil, Colmenarejo, Cabi, Saxton, & Freitas, 2017; Graves et al., 2016). Though these systems vary in detail, each innately incorporates some set of microprocessor-like instructions that are operations over variables. In other domains, like language, many models innately include things like tags for parts of speech; some include innate mechanisms for representing complex, recursive structure.

How long the list really ought to be is unknown. Pinker (1994) for example, proposed a significantly more extensive collection of candidate innate, evolved building blocks for humans, including intuitive mechanics, intuitive biology, number, habitat selection, danger including fear, caution, and phobia, mental maps for large territories, food, contamination, monitoring of well-being, intuitive psychology, a mental Rolodex, self-concept, justice, kinship including elements such as nepotism and parenting, and mating, and concomitants such as sexual attraction and love,

While all of these are empirical claims, each is supported by data drawn from fields such as cognitive and developmental psychology, ethology and neuroscience (Pinker, 1997). Each of his candidate domains appears to universal across humans, and in most instances there is robust evidence for them early in life. (Sex of course develops later, but undeniably partially under genetic control). In addition, neural substrates for most of the candidates on Pinker's list are consistently localized across individuals, suggesting an important degree of genetic contribution to their neural organization

Whether similar machinery is required for artificial general intelligence, it is, of course, an open empirical question. Some of the specifics that Pinker proposed (e.g. concerning mate selection) presumably are not relevant for artificial intelligences, others might perhaps to some degree derive from more general mechanisms (e.g. for cost-benefit analysis). But many others might, as Spelke and Carey (Carey & Spelke, 1994) have argued, start with relatively thin "cores" that are supplemented by cultural and experiential learning.

The important point, for present purposes, is, that there is a whole world of possible innate mechanisms that AI researchers might profitably consider; simply presuming by default it is desirable to include little or no innate machinery seems, at best, close-minded. And, at worst, an unthinking commitment to relearning everything from scratch may be downright foolish, effectively putting each individual AI system in the position of having to recapitulate a large portion of a billions years of evolution.

Incidentally, one of the weakest rejoinders I have heard to the line of argument presented in this article is the dubious claim that evolution is just another form of learning, but over a larger time scale. Well maybe, but that weakens the notion of "learning" immeasurably, such that it encompasses literally everything on either side of the debate, from what Locke had in mind to whatever Plato and Chomsky and Pinker had in mind, and everything in between. Relabeling the debate doesn't resolve the issues, either; one might as well just use the term learning to refer to all change over time, regardless of mechanism, and count rock formations as the product of learning, too.

Evolution (whether through natural selection or simulated artificial techniques) is a means towards building machinery with embedded prior knowledge, not an alternative to prior knowledge.

To be sure, for now, the negative argument in this article is weaker than the stronger argument. The negative argument here starts with a case study of AlphaStar, and the observation that DeepMind has not really taken out as much human knowledge they professed to, and further observation that they are unlikely to achieve the same level of success if they do; these observations seem indisputable.

The positive argument, which is weaker, begins with the speculation that in adding in the right knowledge (Monte Carlo tree search and convolution for board games, convolution without Monte Carlo tree search for the Atari games, and so forth) DeepMind has done better than they might have if they had dropped such knowledge (no convolution or Monte Carlo Search for either Go or Atari). That argument too, seems sound, as far it goes.

What we don't really know is how general *that* argument is -- because we haven't really tried. To my knowledge there has been no known systematic search of what types of (innate) knowledge might be most helpful across a broad array of different types of problems.

In many ways, the current degree of antipathy towards "heavy-handed" a priori knowledge seems like a reversion from what seemed clear in an earlier era. To take one example, although Yann LeCun has recently argued for a strenuously anti-nativist position, his most famous work, on convolution, a widely-adopted, and greatly valuable approach to innately embedding translational invariance in neural networks(LeCun, 1989) was based on the premise that "It is usually accepted that good generalization performance on real-world problems cannot be achieved unless some a priori knowledge about the task is built into the system." LeCun's original conclusion still seems apt: "generalization goes up ... as the amount of built-in knowledge goes up."

Of course, the right final system for AI might or might look like a neural network; such systems have peaked in popularity before, only to decline. But whether neural networks or some other architecture, or combination of architectures (Marcus, 2018) gets us to artificial general intelligence, LeCun's observation still holds.

# 6.  Two methodologies for proceeding in AI

One approach to discerning how much innateness might be required for AI would be to create synthetic agents that do difficult tasks, with some initial degree of innateness, achieve state of the art performance with those tasks, and then iterate, reducing as much innateness as possible, ultimately converging on some (putatively) minimal amount of innate machinery. In the AlphaStar series of papers, DeepMind has essentially followed exactly this strategy.

One might refer to this strategy as a "reductive" strategy, in which the goal is to achieve success with as small (reduced) set of innate machinery as possible. Demis Hassabis's 2017 presentation at NIPS is perhaps the finest example of that reductive strategy I have ever seen.

But is that the right strategy? I see two problems; first, a smaller set of innate primitives is not inherently better; for example, biological organisms may sometimes build in machinery that in principle could be learned, perhaps because there is less risk to an organism if certain information is inherent rather than hard won through individual, potentially life-threatening example. A robot could learn, through experience, how to walk, but there could be advantages in having it come from the factory already capable of walking. If you are baby ibex scaling the side of cliff, you may be better off with a small but focused set of innate priors than with a more plastic system that would require a large number of life-threatening experiences. Or, as is so often the case in biology, with the mixture; our motor systems may allow us step from birth when submerged partly in water. And then that innate prior for taking alternating steps gets calibrated with experience. Such a split system may be more robust than one that learns in any entirely unconstrained fashion. As Brad Wyble noted in an email, "evolution clearly has enormous flexibility in specifying the level of nativism for a given organism that is appropriate to the demands placed on it from birth. The question is never nativism or not, but always: how much is a optimal for a given case?"

Second, there is a methodological issue. If a reductive strategy succeeds in finding a small set of innate machinery that suffices for a narrow category of tasks, you cannot automatically conclude that the same machinery will suffice for all tasks. All you can reasonably conclude is that this machinery may well suffice for other tasks in this category, with diminishing effectiveness the further one moves from its core cases. The success of AlphaZero on Go, chess, and shoji, suggests that its mechanisms may well suffice to achieve super-human performance on perfect knowledge, two-player, zero-sum, deterministic, discrete games. These mechanisms may not suffice even for some other games, let alone for cognitive tasks generally.

An alternative approach might start from the top down, examining properties of (e.g.) adult or child cognitive systems, seeking clues as to what might be innate (based on behavioral grounds), or conceptually necessary (as Kant argued with respect to space and time, in his *Critique of Pure Reason* (Kant, 1781). Chomsky's argument for tree structure extended from empirical considerations (through the study of multiple languages); my own arguments [in The Algebraic

Mind] stemmed primarily from empirical considerations about the nature of human cognition. Spelke's view appears to come from both directions, empirical (from the study of human infants) and conceptual, with an emphasis on six systems, quoting this lucid distillation from a recent email:

> *What I believe [empirical experiments] have told us is that there are (at least) 6 systems of core knowledge, each centering on a set of interconnected, abstract variables: (1) objects & their motions, though it isn't clear yet whether the abstract variables are dynamic, like forces and masses, or kinematic, like bodies and motions. (2) agents & their intentions & actions (the interconnected variables include those of utility theory ... and notions of perceptual access, but no notion of communication, cooperation, sharing, or phenomenal mental states; (3) social beings & their states of engagement (this system gives rise both to intuitive sociology and to the intuition that other people (and pets, etc) have phenomenal states like ours; (4) number (the approximate number system); (5) geometry for navigation (distance & direction); and (6) geometry for object recognition .. evolved to recognize and categorize living kinds—and then hijacked for recognizing artifacts as we started to invent them. (There may be more systems but I think not too many more, because with the proliferation of core systems comes a search problem: the infant will need more & more information to figure out which system to use on any particular occasion.*

The reductive approach is about distilling a set of cognitive primitives, by successively factoring out needless complexity, in the fashion that Hassabis described at NIPS. The topdown approach is about using what we independently know about cognition in order constrain what we might as starting points for AI.

Neither approach, reductive or topdown, is inherently superior; there is virtue in the reductive approach followed in the AlphaStar work, and in the topdown approach exemplified by Chomsky, Pinker, Spelke, and myself.[7]

Both approaches have their advantages. The point of the current paper is that, on the one hand, the matter is hardly settled, on the other, the balance between the two approaches has, across the field of machine learning, become seriously distorted. It's time for AI to take nativism more seriously.

---

[7] DeepMind itself uses both approaches, from time to time, although not so much in the AlphaStar project; my issue in this paper is not with DeepMind's research program as a whole, but solely with the empiricist interpretation that they gave the work.